\documentclass[sigconf]{acmart}
\usepackage{amsmath}        
\usepackage{graphicx}       %
\usepackage[ruled,linesnumbered,longend]{algorithm2e}   
\usepackage{multirow}
\usepackage{makecell}

\AtBeginDocument{%
  \providecommand\BibTeX{{%
    \normalfont B\kern-0.5em{\scshape i\kern-0.25em b}\kern-0.8em\TeX}}}

\begin{document}

\title{Towards Trustworthy Unsupervised Domain Adaptation: A Representation Learning Perspective for Enhancing Robustness, Discrimination, and Generalization}



\author{Jia-Li Yin, Haoyuan Zheng, Ximeng Liu}
\email{jlyin@fzu.edu.com, hyzheng@fzu.edu.com,snbnix@gmail.com}
\affiliation{%
  \institution{College of Computer Science and Big Data, Fuzhou University}
  \city{Fuzhou}
  \country{China}}


\begin{abstract}

  Robust Unsupervised Domain Adaptation (RoUDA) aims to achieve not only clean but also robust cross-domain knowledge transfer from a labeled source domain to an unlabeled target domain. A number of works have been conducted by directly injecting adversarial training (AT) in UDA based on the self-training pipeline and then aiming to generate better adversarial examples (AEs) for AT. Despite the remarkable progress, these methods only focus on finding stronger AEs but neglect how to better learn from these AEs, thus leading to unsatisfied results. In this paper, we investigate robust UDA from a representation learning perspective and design a novel algorithm by utilizing the mutual information theory, dubbed MIRoUDA. Specifically, through mutual information optimization, MIRoUDA is designed to achieve three characteristics that are highly expected in robust UDA, \textit{i.e.}, robustness, discrimination, and generalization. We then propose a dual-model framework accordingly for robust UDA learning. 
  Extensive experiments on various benchmarks verify the effectiveness of the proposed MIRoUDA, in which our method surpasses the state-of-the-arts by a large margin.
  
\end{abstract}





\maketitle

\section{Introduction}
\label{sec:1}

\begin{figure}
  \centering
  \includegraphics[width=1.0\linewidth]{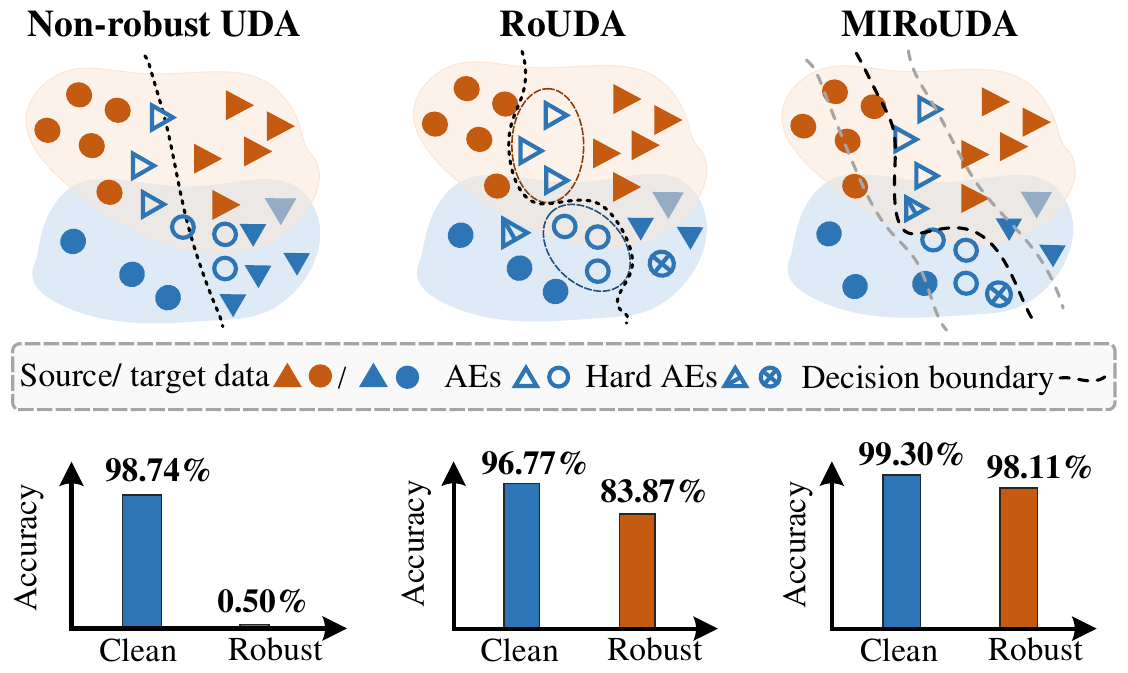}
  \caption{Different UDA schemes and their performance. Left: Traditional UDA methods do not take robustness into account, resulting in defenseless against adversarial attacks. Middle: Existing RoUDA can improve the robustness via incorporating AT but the results are under-optimal due to their empirical design. Right: Our MIRoUDA unifies the MI theory for improving robustness, discrimination, and generalization. We use CDAN~\cite{long2018conditional} as the UDA baseline, and PGD-20~\cite{madry2018towards} for evaluating model robustness. The results of RoUDA are from the SRoUDA~\cite{zhu2022srouda} proposed in 2023.} 
  \label{fig:intro}
\end{figure}

Deep neural networks (DNNs) have brought various applications to a new era with performance pulled ahead of human-like accuracy. However, these advances come only when a large amount of labeled training data is available. Constrained by the expensive labeled data acquisition in the real world, unsupervised domain adaptation (UDA) \cite{chen2019transferability,ganin2015unsupervised,long2015learning,long2018conditional,sdat_rangwani2022closer,daln_chen2022reusing,DisClusterDA,covi_na2022contrastive,fixbi_na2021fixbi,DARE_GRAM_Nejjar_2023_CVPR,COTUDA_Liu_2023_CVPR}, which aims to transfer the knowledge learned from a labeled source domain to an unlabeled target domain, has been widely developed to adapt for real-world applications. In practice, however, we are facing more problems in the deployment of DNNs. Recent studies~\cite{madry2018towards,tack2022consistency,wang2020improving,xie2019feature,zhang2019theoretically, CFA_Wei_2023_CVPR,Enemy_Dong_2023_CVPR} have reported the vulnerability of DNNs to adversarial attacks, \textit{i.e.}, adding imperceptible noise into benign data can cause dramatic changes in DNN predictions, which arises severe trustworthy concerns. Taking the UDA task \textbf{D} $\rightarrow$ \textbf{W} as a showcase in Figure\,\ref{fig:intro}, typical UDA methods would minimize the representation discrepancy between the source and target data without considering robustness, resulting in a high clean accuracy but $0.50\%$ robust accuracy against adversarial attacks. 

As a feasible solution, RoUDA aims to improve the model robustness against adversarial attacks while achieving domain adaptation. Commonly, a UDA model is considered robust when the model can prefer robust feature representations during learning so that it can defend against the adversarial attacks. However, it would lead to a trivial solution if the robustness is over-emphasized, \textit{i.e.}, the features are invariable. Hence, in addition to robustness, discrimination and generalization are also highly expected in RoUDA. Specifically, the discrimination can avoid the aforementioned trivial solution, and the generalization can help the robust generalization to other unseen attacks, which can significantly benefit the practical applications.

To achieve RoUDA, a number of works have been conducted to explore how to inject robustness into UDA. Their main differences lie in the robust source from either distillation from external robust models~\cite{awais2021adversarial} or directly incorporating adversarial training (AT)~\cite{lo2022exploring,zhu2022srouda,10.1145/3503161.3548323} into UDA. In brief,~\cite{awais2021adversarial} employs a pre-trained robust model to distill the robust knowledge during the UDA process. Although model robustness can be achieved by this method, the distillation performance is sensitive to the architecture of the external model. Another line of methods~\cite{lo2022exploring,zhu2022srouda,10.1145/3503161.3548323} proposes performing AT based on the self-training pipeline, where a source model is first pre-trained on the labeled source data and then the robust target model can be obtained by applying AT on the target data with pseudo labels produced from the pre-trained source model. Such methods directly inject AT into UDA process and can effectively improve model robustness. However, almost all of them only seek to generate better pseudo labels for applying AT, which leads to under-optimal results, where some hard AEs are difficult to classify as shown in the middle part of Figure\,\ref{fig:intro}. On one hand, noisy labels are inevitable in the self-training pipeline; on the other hand, existing methods generally neglect how to learn from these AEs, which is also crucial for improving model robustness and retaining clean accuracy.

In this work, inspired by the success of the previous self-training-based RoUDA methods, we keep along this line but from a different perspective. It has been pointed out recently that slight label noise may not harm AT and can even improve robust generalization~\cite{zhang2022noilin,Jingfeng_noilin,zhu2022srouda}. Thus, we choose not to struggle for more accurate pseudo labels for target data, instead, we take the perspective of representation learning and utilize mutual information (MI) theory~\cite{Zeng_2023_CVPR,Zhang_2023_CVPR} to encourage robust, discriminative, and generalized learning during the self-training-based UDA process. Specifically, we first theoretically show that robustness could be formulated as the minimization of the MI $I(X;F)$ between input $X$ and feature representations $F$. Meanwhile, we prove that discrimination could be achieved by maximizing $I(F;Y)$ between the feature representations and corresponding labels $Y$. In this case, robust and discriminative features are preferred during model learning and thus RoUDA could be achieved. To further improve the model generalization to unseen attacks, we develop a dual-model architecture to learn diverse representations and we show that the generalization can be achieved by minimizing the MI $I(F_1;F_2)$ between the representations from the dual models. Additionally, to further mitigate the clean accuracy drop brought by the noisy pseudo labels, we equip a consensus regularizer to force the output consensus between the source and target model on clean target data during target model training. The proposed method, dubbed MIRoUDA, demonstrates superior performance in extensive experiments, where the model robustness is improved by a large percentage ($0.50\% \xrightarrow{} 98.11\%$ in Figure\,\ref{fig:intro}). To summarize, our contributions are as follows:

\begin{itemize}
    \item We unify the RoUDA task under the MI theory and propose a novel theoretical-grounded MI based RoUDA method, dubbed MIRoUDA, for achieving robustness, discrimination, and generalization representation learning during UDA process.

    \item We theoretically show that the robustness and discrimination could be achieved by minimizing MI between inputs and feature representations while maximizing MI between representations and labels. Moreover, the generalization could be achieved by minimizing MI between the representations of two diverse models. 

    \item Driven by the unified MI theory, we propose a dual-model architecture accordingly for RoUDA learning. With the help of an additional consensus regularizer, the proposed MIRoUDA achieves superior performance on various benchmarks in terms of both clean and robust accuracy.
\end{itemize}

\begin{figure*}[t]
  \centering
  \includegraphics[width=0.95\textwidth]{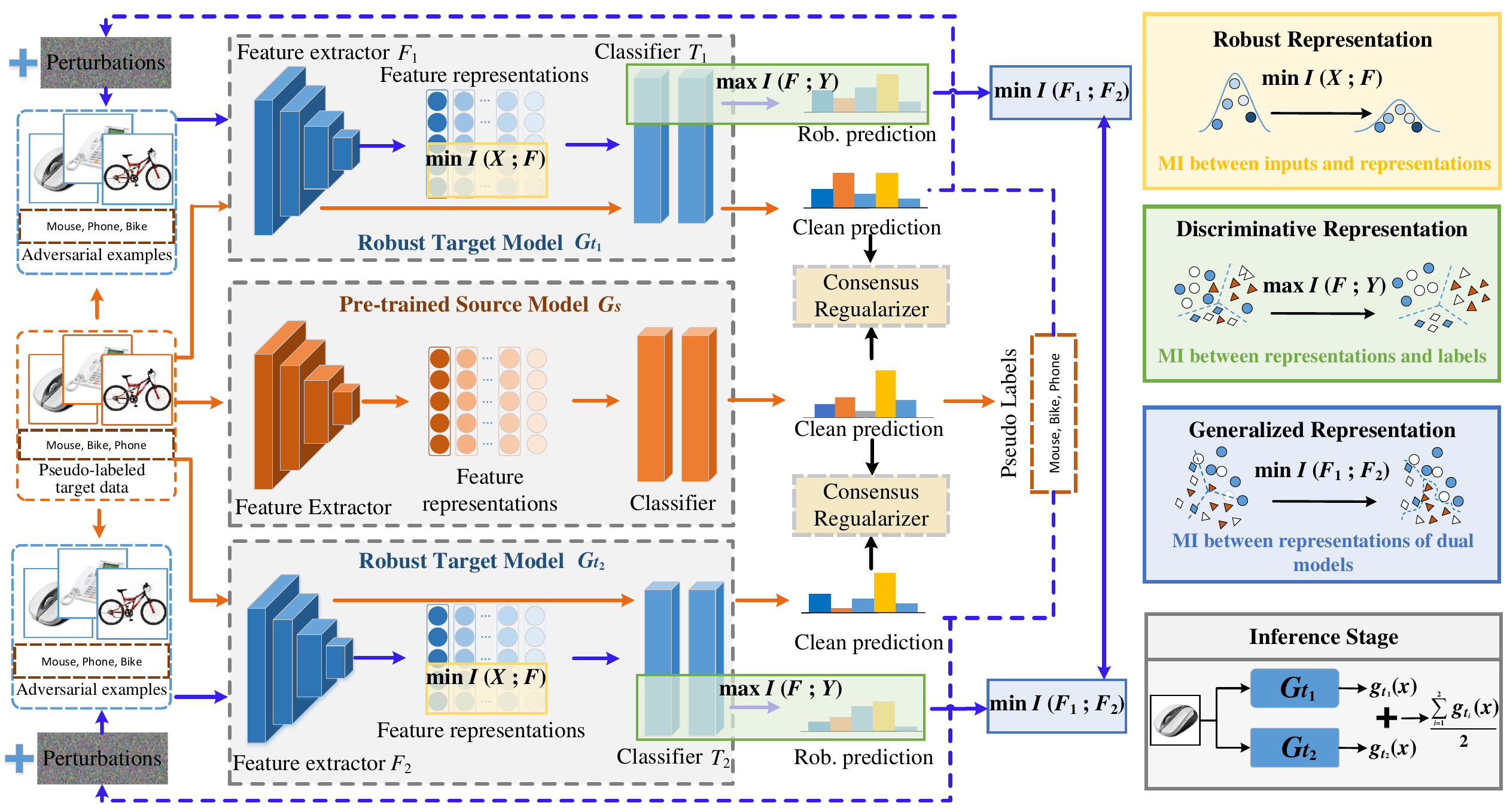}
  \caption{Overview of our proposed MIRoUDA. The whole framework is based on the self-training pipeline, where a source model $G_s$ is first pre-trained by the UDA baseline. We omit this part due to its commonplace. Here we put emphasis on the robust target training part, where two dual models are trained parallelly for MI optimization. We use the MI objective $\min I(X;F)$, $\max I(F;Y)$, and $\min I(F_1;F_2)$ to achieve robust, discriminative, and generalized representation learning, respectively. A consensus regularizer is equipped to force the output consensus between the source and robust model on clean target data, which consequently helps in preserving high clean accuracy.}
  \label{fig:overview}
\end{figure*}

\section{Related Work}
\noindent\textbf{Unsupervised Domain Adaptation.}
UDA has been an active research field in recent years and numerous works have been proposed. Early strategies utilized a specific metric to reduce the distribution divergence, such as Maximum Mean Discrepancy (MMD) in DAN~\cite{long2015learning} and JAN~\cite{long2017deep}. Inspired by adversarial learning, DANN~\cite{ganin2015unsupervised}, CDAN~\cite{long2018conditional}, BSP~\cite{chen2019transferability} and SDAT~\cite{sdat_rangwani2022closer} added an additional domain discriminator to force more discriminative domain-invariant features. Recently, another line of works~\cite{Xu_2019_ICCV_ST,Choi2019PseudoLabelingCF,Yang2022ST3D,Fan_2022_CVPR} explored self-training scheme to generate pseudo labels for the target domain and then re-train the model by pseudo-labeled target data. To improve the quality of pseudo labels, efforts are devoted to reducing label noise by utilizing progressive generation strategy~\cite{Xu_2019_ICCV_ST}, curriculum learning~\cite{Choi2019PseudoLabelingCF}, and voting schemes~\cite{Yang2022ST3D,Fan_2022_CVPR}. Although remarkable performance has been achieved by the aforementioned works, they generally only focus on improving the accuracy of UDA models and neglect the model robustness.
\vspace{5pt}

\noindent\textbf{Adversarial Attacks and Defenses.}
Since Szegedy \textit{et al}. first revealed the vulnerability of DNNs against imperceptible perturbations in \cite{szegedy2013intriguing}, the adversarial robustness of models has attracted increasing attention and triggered two research directions, namely \textit{adversarial attacks and defenses}, as the former one trying to develop powerful attack strategies for misleading models and the latter one aiming to improve the model robustness against these attacks. For the attacks, typical algorithms utilize the model gradient descent to generate the perturbations including Fast Gradient Sign Method (FGSM)~\cite{goodfellow2014explaining}, Projected Gradient Descent (PGD)~\cite{madry2018towards}, and AutoAttack (AA)~\cite{AA_Croce}. For the defenses, numerous methods have been proposed for improving the model robustness, among which adversarial training (AT)~\cite{madry2018towards} has been proven to be the most effective defense method. The core idea of AT is to leverage online-generated adversarial examples in the training set so that the model can prefer more robust representations during learning. Modifications are developed to further improve the robustness in AT, with changes in the adversarial examples generation procedure~\cite{tramer2017ensemble,58kannan2018adversarial,zheng2020efficient}, model parameter updating~\cite{Hwang2021SWA,robust-tpami} and feature adaption~\cite{wang2021adaptive,Yin2023}.
\vspace{5pt}

\noindent\textbf{Adversarial Robustness of UDA Models.}
In contrast to extensive UDA research working on improving model accuracy on clean target data, few have explored the adversarial robustness of UDA models. Based on how robustness is incorporated, the existing works could be roughly divided into two categories, namely, robust-distillation-based methods and AT-based methods. In brief, the robust-distillation-based methods utilize another robust model to distill the robust knowledge during the UDA process. Specifically, Awais \textit{et al}.~\cite{awais2021adversarial} proposed robust feature adaptation (RFA) where an external adversarially pre-trained model was used as a teacher model to distill robustness. Although it is effective in improving the robustness, this approach is quite sensitive to the architecture and perturbation budget of the teacher model. Different from the distillation methods, the AT-based methods combine AT in the UDA process. However, AT needs a supervised loss function to generate the adversarial examples for training. To solve this problem, Lo \textit{et al}.~\cite{lo2022exploring} presented ARTUDA that approximates the adversarial examples through an unsupervised loss function. Unfortunately, such adversarial examples cannot guarantee the inner-maximization in AT thus leading to unsatisfactory robustness.

Motivated by the success of the self-training pipeline in UDA~\cite{Xu_2019_ICCV_ST,Choi2019PseudoLabelingCF,Yang2022ST3D,Fan_2022_CVPR} which generates pseudo labels for target data and allows the direct inject of AT, some studies have been carried out on self-training pipeline and use pseudo labels for AT. For example, Yang \textit{et al}.~\cite{Yang2021iccv} utilized the naive self-training pipeline for robust image segmentation. Zhu \textit{et al}.~\cite{zhu2022srouda} make a step forward to explore better pseudo labels for adversarial target model training by incorporating meta learning. Although promising results have been achieved by the two methods, they generally focus on generating more proper or stronger adversarial examples but neglect how to learn from these adversarial examples in AT. Moreover, both of them are designed heuristically and lack theoretical explanations or supports. In this paper, we keep along the self-training based pipeline for RoUDA and explore how to learn adversarial examples from a representation-learning perspective for better robustness, discrimination, and generalization.

\section{Method}

In this section, we first give the mathematical definition of RoUDA. We then elaborate on how to learn robust, discriminative, and generalized feature representations through a unified information theory and propose a dual-model framework for RoUDA learning accordingly. 

\subsection{Probelm Definition}
Given a labeled source domain dataset \(\mathcal{S}=\{(x^s_i,y^s_i)\}^N_{i=1}\) and an unlabeled target domain dataset \(\mathcal{T}=\{(x^t_i)\}^N_{i=1}\). The goal of UDA is to train a classifier $\mathbf{G}$, composed of a feature extractor $\mathbf{F}$ followed by a classifier $\mathbf{T}$, 
that can predict correctly on the unlabeled target data. The RoUDA takes a further step, where not only clean accuracy is achieved but also robust to adversarial attacks. Based on this, we mathematically formulate RoUDA as follows:
\vspace{2pt}

\noindent\textbf{Definition 1 \textit{Robust UDA}.} \textit{Given a classifier $\mathbf{G}$, let $x_i^t$ and $\hat{x}_i^t$ be the target domain samples and their corresponding adversarial examples, the UDA model is robust if}
\begin{equation}
    \label{eq:ruda}
    \mathbf{G}({x}_i^t)=\mathbf{G}(\hat{x}_i^t), 
\end{equation}
\textit{where $\hat{x}_i^t$ is typically generated by adding a perturbation on the original image $x_i^t$, where PGD-$k$ attack~\cite{madry2018towards} is usually applied in an iterative way as follows:}
\begin{equation}
    \label{eq:ae}
    \hat{x}_{i,k+1}^t= \Pi_{\mathcal{B}(x_i^t)}(\hat{x}_{i,k}^t+\alpha\cdot sign(\nabla_{\hat{x}_{i,k}^t} \ell_{ce}(G(\hat{x}_{i,k}^t), y_i^t)),
\end{equation}
\textit{where $\hat{x}_{i,k}^t$ is initialized as the input $x_i^t$, and the final adversarial example $\hat{x}_i^t = \hat{x}_{i,k_{max}}^t$, where $k_{max}$ is the maximum number of iterations; $\mathcal{B}(x_i^t)$ denotes the $\ell_{p}$-norm ball centered at $x_i^t$ with radius $\epsilon$, i.e., $\mathcal{B} (x_i^t)=\{\hat{x}_i^t:\|\hat{x}_i^t-x_i^t\|_{p} \leq \epsilon\}$; and $\Pi$ refers to the projection operation for projecting the adversarial examples back to the norm-ball.}

Notably, robustness could be trivially achieved by encouraging invariant representation learning, but it is obviously not a reasonable solution. In addition to robustness, a good RoUDA method is also expected to embrace higher discrimination and generalization. Specifically, a discriminative UDA method can avoid the aforementioned trivial solution brought by the over-emphasized robustness, and the generalization can help the model defend against various unseen attacks in practical scenarios. To achieve these goals, we propose to utilize the mutual information theory and propose a theoretical-grounded dual-model framework accordingly, which we describe in what follows.

\subsection{RoUDA via Mutual Information Optimization}

As stated above, in this paper, we achieve RoUDA by pursuing three objectives, namely robust, discriminative, and generalized representation learning, which are daunting to formulate and optimize jointly. Here we first theoretically show that these three diverse objectives could be derived from a unified theory, \textit{i.e.}, the optimization of MI during model learning. Specifically, robustness could be achieved by minimizing the MI between adversarial inputs ${X}$ and feature representations ${F}$, \textit{i.e.}, $I({X};{F})$; discrimination could be obtained by the maximization of the MI $I({F};{Y})$ between feature representations and labels; and the generalization could be learned by minimizing the MI between the feature presentations produced by two dual models $I({F_1};{F_2})$. In the following, we will present the mathematical details for the mutual information optimization and the proposed dual-model framework in turn.

\subsubsection{Learning robust representation via $\min I(X;F)$}
\label{sec:min_xf}
We begin with the discussion about the relationship between mutual information $I({X};{F})$ and robust representation. We prove why minimizing $I({X};{F})$ can make the learned representation more robust and then make it tractable in the training process. First, we mathematically formulate robust representation as follows:

\noindent\textbf{Definition 2 \textit{Robust representation}.} \textit{Given unlabeled data $x^t$ from the target domain and a feature extractor $\mathbf{F}$, the learned representation is robust if}
\begin{equation}
    \label{eq:robust_representation}
    \forall_{A \sim \mathcal{A}} || \mathbf{F}(x^t)-\mathbf{F}(A(x^t))||_p < \epsilon,
\end{equation}
\textit{where $A$ is an adversary transformation from $\mathcal{A}$, which refers to adversarial attacks here. $\epsilon$ is the error factor which is positive and close to zero.}

Clearly, if the distance between the features $\mathbf{F}(A(x^t))$ generated from an adversarial example and the features $\mathbf{F}(x^t)$ extracted by original inputs $x^t$ is less than $\epsilon$ under the $l_p$-norm, it indicates that the learned representation is more robust to adversarial perturbations. In information theory~\cite{PhysRevE.69.066138,pmlr-v80-belghazi18a,alemi2017deep,Zhang_2023_CVPR}, the mutual information $I(X;F)$ between inputs and feature representations could quantify the independence between random variables. Thus, we can formulate the objective function as:

\begin{equation}\label{eq:xffth}
\begin{aligned}
    \min & I(X;F)=H(F)-H(F|X)\\
    &=-\sum \limits_{i=1} ^{N} p_{f_i}\log p_{f_i}-(-\sum \limits_{i=1}^{N} p_{f,\hat{x}^t_i}\log p_{f|\hat{x}^t_i}),
 \end{aligned}
\end{equation}
where $H(\cdot)$ denotes the entropy, $p_{f_i}$ is the marginal probability density of feature representations, and $p_{f|\hat{x}^t_i}$ refers to the conditional probability density given the $i$-th adversarial example $\hat{x}^t_i$. The minimization of the first term $H(F)$ encourages general feature representations over the inputs, and the maximization of the second term $H(F|X)$ pushes invariable feature representations given certain adversarial examples, which all lead to \textbf{robust} feature representations.

Next, we further make $I(X;F)$ tractable in the training procedure. From Eq.\,\eqref{eq:xffth}, due to the marginal density function $p_{f_i}$ is not accessible, we alternatively estimate it with the variational distribution $q(f_i)$. According to the non-negativity of KL divergence, we have $\int p_{f_i}log(p_{f_i})\geq\int q(f_i)log(q(f_i))$, and then we can optimize Eq.\eqref{eq:xffth} by estimating the upper bound:
\begin{equation}
   \begin{aligned}
        I(X;F) 
        &\leq \int_{X}p(X)\int_{F}p(F|X)log\frac{p(F|X)}{q(F)}\\
        &=\int_{X}p(X) D_{KL}(p(F|X)||q(F)). 
    \end{aligned}
\end{equation}

Assuming it obeys the Gaussian distribution, \textit{i.e.},
$p(F|X)\sim N(\mu(F(\hat{x}^t)),\sigma^2 I)$ and $q(F)\sim N(0,I)$. Then we have:
\begin{equation}
   \begin{aligned}
        D_{KL} & (N(\mu,\sigma^2)||N(0,I))\\
        &=\int\frac{1}{\sqrt{2\pi}\sigma}e^{-\frac{(x-\mu)^2}{2\sigma^2}}
        \log\frac{\frac{1}{\sqrt{2\pi}\sigma}e^{-\frac{(x-\mu)^2}{2\sigma^2}}}{\frac{1}{\sqrt{2\pi}\sigma}e^{-\frac{x^2}{2\sigma^2}}}dx \\
        &=\int\frac{1}{\sqrt{2\pi}\sigma}e^{-\frac{(x-\mu)^2}{2\sigma^2}}
        \log\frac{1}{\sigma}e^{\frac{1}{2}(x^2-\frac{(x-\mu)^2}{\sigma^2})}dx \\
        &=\int\frac{1}{\sqrt{2\pi}\sigma}e^{-\frac{(x-\mu)^2}{2\sigma^2}}
        (-\log\sigma+\frac{1}{2}(x^2-\frac{(x-\mu)^2}{2\sigma^2}))dx.
    \end{aligned}
\end{equation}

The first term is $-\log\sigma$, the second term is $\frac{(\mu^2+\sigma^2)}{2}$, and the third term is $-\frac{1}{2}$. As $\sigma$ is a constant, then the result can be represented as:
\begin{equation}
    \label{eq:min_xf}
    I(X;F) \leq \frac{1}{2}||\mu(\mathbf{F}(\hat{x}^t))||_2^2+C,
\end{equation}
where $C$ is a constant that can be ignored. Minimizing $I(X;F)$ is equivalent to applying the $l_2$-norm regularization to the features, which can be written as:
\begin{equation}\label{eq:xfloss}
\begin{aligned}
    \mathcal{L}_{rob}=||\mu(\mathbf{F}(\hat{x}^t))||_2^2.
 \end{aligned}
\end{equation}

\subsubsection{Learning discriminative representation via \texorpdfstring{$\max I(F;Y)$}{I(F;G)}}
\label{sec:max_fg}
Here we reveal that the discriminative representation could be learned via maximizing $I(F;Y)$ between the features and the corresponding labels. Without loss of generality, we denote ${F}$ as the features of adversarial examples from target data $x^t$ and denote ${Y}$ as the corresponding labels. Note that the labels are generated from a pre-trained source model. Then, we can formalize $I(F;Y)$ as follows:
\begin{equation}\label{eq:FY}
\begin{aligned}
    \max  I(F;Y)&=H(Y)-H(Y|F),
 \end{aligned}
\end{equation}
where $H({Y})$ is a constant so that it is equivalent to the minimization of the conditional entropy $H({Y}|{F})$, which can be represented as:
\begin{equation}\label{eq:FYy}
    H(Y|F)=-\sum \limits_{i=1}^{N} p_{y^t_i,F(\hat{x}^t_i)}\log p_{y^t_i|F(\hat{x}^t_i)},
\end{equation}
where $p_{y^t_i|F(\hat{x}^t_i)}$ is the conditional probability of the feature of sample $\hat{x}^t_i$ being assigned to the label $y^t_i$. Thus, the minimization of $H({Y}|{F})$ encourages feature representation of each sample to its corresponding class and away from others, \textit{i.e.}, the model is encouraged to produce \textbf{discriminative} feature representations. In the training procedure, we can achieve $\max I({F};{Y})$ by minimizing the softmax-cross entropy loss as follows:
\begin{equation}\label{eq:FYloss}
\begin{aligned}
    \mathcal{L}_{dis}=\ell_{ce}(\mathbf{G}(\hat{x}^t),y^t).
 \end{aligned}
\end{equation}

Note that solely optimizing the learned representation with $\min I(X;F)$ would prefer robust feature representations and not lead to discrimination. As a result, it is necessary to combine both $\min I(X;F)$ and $\max I(F;Y)$ to achieve both accurate prediction and robust representation.

\begin{table*}
    \centering
    \caption{Comparison of clean and robust accuracy \% of different methods on \textbf{Office-31} dataset. Note that $\dag$ denotes the results copied from their original papers, and $-$ denotes that the results are not reported.}
    \label{tab:office-31}
    \resizebox{1.0\textwidth}{!}{
        \begin{tabular}{ccccccccccccccc}
            \toprule
            \multirow{2}{*}{Methods} 
            & \multicolumn{2}{c}{\textbf{A} \(\xrightarrow{}\) \textbf{W}} & \multicolumn{2}{c}{\textbf{W} \(\xrightarrow{}\) \textbf{A}} 
            & \multicolumn{2}{c}{\textbf{A} \(\xrightarrow{}\) \textbf{D}} & \multicolumn{2}{c}{\textbf{D} \(\xrightarrow{}\) \textbf{A}}
            & \multicolumn{2}{c}{\textbf{W} \(\xrightarrow{}\) \textbf{D}} & \multicolumn{2}{c}{\textbf{D} \(\xrightarrow{}\) \textbf{W}}
            & \multicolumn{2}{c}{\textbf{Avg.}}\\
             
            \cmidrule{2-15} 
            &Clean&Rob. &Clean&Rob. &Clean&Rob. &Clean&Rob. &Clean&Rob. &Clean&Rob. &Clean&Rob.\\
            \midrule
            
            Baseline(UDA) & 91.19 & 0.00  & \textbf{73.30} & 11.96 & 92.97 & 0.00  & \textbf{73.37} & 6.82  & \textbf{100}   & 0.00  & 98.74 & 0.50  & 88.26 & 3.21  \\ \midrule
            Source-only   & 26.72 &7.45   &15.36  &4.97  &18.23  &8.12 &7.29 &6.53 &65.32 &25.89 &43.26 &27.51 &29.36 &13.41 \\ \midrule
            AT+UDA        & 65.16 & 30.82 & 54.49 & 34.93 & 62.85 & 16.87 & 33.55 & 22.58 & 96.39 & 66.67 & 85.28 & 68.43 & 66.29 & 40.05 \\ \midrule
            UDA+AT        & 88.67 & 76.73 & 69.33 & 60.28 & 85.00 & 58.00 & 68.97 & 63.12 & 89.00 & 55.00 & 94.34 & 78.62 & 82.55 & 62.59 \\ \midrule
            ARTUDA\cite{lo2022exploring} \dag           & - & - & - & - & - & - & - & - & - & - & 95.20 & 92.50 & - & - \\ \midrule
            RFA\cite{awais2021adversarial} \dag         & - & - & - & - & - & - & - & - & - & - &- & - & 84.21 & 74.31 \\ \midrule
            AFSR\cite{10.1145/3503161.3548323}  \dag         & 89.10 & 81.00 & 68.30 & 60.30 & 86.4 & 67.60 & 68.50 & 59.00 & 99.4 & 96.00 & 99.30 & 93.80 & 85.20 & 76.30 \\ \midrule
            SRoUDA\cite{zhu2022srouda}    \dag    & \textbf{95.97} & 84.68 & 67.10 & 57.42 & 91.94 & 85.48 & 72.47 & 57.20 & {100}   & 88.71 & 96.77 & 83.87 & 87.27 & 75.79 \\ \midrule
            Ours          & 94.34 & \textbf{91.19} & 72.16 & \textbf{67.20} & \textbf{93.00} & \textbf{83.00} & 72.81 & \textbf{67.20} & \textbf{100} & \textbf{99.00} & \textbf{99.30} & \textbf{98.11} & \textbf{88.60} & \textbf{84.28} \\ 
            \bottomrule
        \end{tabular}
    }
    
\end{table*}

\subsubsection{Learning generalized representation via \texorpdfstring{$\min I({F_1};{F_2})$}{I(F;F')}}
To further improve the model generalization, we develop a dual-model-based architecture to learn diverse representations where two sub-models are trained parallelly. Given two models ${\mathbf{G}_1}, {\mathbf{G}_2}$ including feature extractor ${\mathbf{F}_1}$ and ${\mathbf{F}_2}$, we focus on the generalization that contributes to making the learned features defend against unforeseen adversaries. It can be achieved by minimizing $I(F_1;F_2)$, which encourages representation learned by different models to be independent of each other. To be specific, given $I(F_1;F_2)=H(F_1)-H(F_1|F_2)$, the $\max I(F_1;F_2)$ is equivalent to the minimization of the conditional entropy $H(F_1|F_2)$ $=-\mathbb{E}_{p({F_1},{F_2})}[\log p(F_1|F_2)]$. Then we have:
\begin{equation}
   \begin{aligned}
        &-\mathbb{E}_{p({F_1},{F_2})} [\log p(F_1|F_2)]\\
        &=-\mathbb{E}_{p({F_1},{F_2})} [\log q(F_1|F_2)]-D_{KL}(p(F_1|F_2)||q(F_1|F_2))\\
        &\leq-\mathbb{E}_{p({F_1},{F_2})} [\log q(F_1|F_2)],
   \end{aligned}
\end{equation}
without loss of generality, we assume $q(F_1|F_2)$ obeys Gaussian distribution $N(F_1;F_2,\sigma^2I)$, then we have $-\mathbb{E}_{p({F_1},{F_2})}[\log q(F_1|F_2)]\propto \mathbb{E}_{p(F_1,F_2)}[||F_1-F_2||^2]+C$, where $C$ is the constant. 
Hence, we can optimize the minimization of $I(F_1;F_2)$ by its negative form:
\begin{equation}
    \label{eq:min_ff}
    \min I({F_1};{F_2}) \propto -\mathbb{E}_{p({F_1},{F_2})} ||{F_1}-{F_2}||_2^2+C.
\end{equation}

Finally, the loss for generalization is then formulated as:
\begin{equation}\label{eq:FFloss}
\begin{aligned}
    \mathcal{L}_{gen}=-||\mathbf{F}_1(\hat{x}_1^t)-\mathbf{F}_2(\hat{x}_2^t)||_2^2,
 \end{aligned}
\end{equation}
where $\hat{x}_1^t$ and $\hat{x}_2^t$ are the adversarial examples generated from the sub models $\mathbf{G}_1$ and $\mathbf{G}_2$, respectively. 
Note that we aim to improve the generalization ability via representation rather than simply maximizing the output difference in two models, which could potentially lead to a decrease in clean accuracy. 

To summarize, we unify the target model learning in the self-training-based RoUDA task from the perspective of information theory. With the above theoretical analysis, we arrive at our loss by combing Eq. \eqref{eq:xfloss}, \eqref{eq:FYloss} and \eqref{eq:FFloss}, \textit{i.e.},
\begin{equation}
    \label{eq:rmi}
    \mathcal{L}_\textbf{MI} = \mathcal{L}_{dis}+\alpha \mathcal{L}_{rob}+ \beta \mathcal{L}_{gen},
\end{equation}
where $\alpha$ and $\beta$ are the hyper-parameters for the trade-off, which we will further discuss in Sec.\,\ref{sec:analysis}.

\subsection{Framework and Training Objective}
\noindent\textbf{Framework.} An overview of the proposed MIRoUDA framework is shown in Figure\,\ref{fig:overview}. Our framework is based on the self-training pipeline, where a source model $\mathbf{G}_s$ is first pre-trained by applying UDA baseline on the benign labeled source data and unlabeled target data, and then the target model $\mathbf{G}_t$ is adversarially trained on the target data with pseudo labels produced from $\mathbf{G}_s$. To improve the robust generalization ability of the target model, we propose a dual-model architecture for the robust model learning. Specifically, there are two parallel subnetworks $\mathbf{G}_{t_1}$ and $\mathbf{G}_{t_2}$ with the same architecture to learn diverse representations, and we apply the MI optimization on both models during training. In the inference phase, the outputs from the two models are averaged to be the final prediction.

\vspace{3pt}
\noindent\textbf{Consistency Regularizer.} Although the proposed MI optimization can significantly improve the robustness, discrimination and generalization of the model, the degeneration of clean accuracy resulting from naive pseudo-label adversarial training still exists. Here apart from $\mathcal{L}_\textbf{MI}$, we additionally equip a simple yet powerful regularizer to guide the model training. The key idea is to force the robust target model to have a consensus logits output with the clean source model on clean target data, thus the robust target model can extract more transferable features and maintain high accuracy on clean data. Here we adopt the JS-Divergence to measure the clean logits divergence between the source and target model, and the consensus regularizer can be described as:
\begin{equation}
    \label{eq:js}
    \mathcal{L}_{cs} = \text{JSD}(\mathbf{G}_s(x^t), \mathbf{G}_t(x^t)).
\end{equation}

\noindent\textbf{Training Objective.} As a result, the overall training objective for our MIRoUDA can be formulated as:
\begin{equation}
    \label{eq:loss}
    \mathcal{L} = \mathcal{L}_\textbf{MI}+\gamma \mathcal{L}_{cs},
\end{equation}
where $\gamma$ is a weighting factor which can be empirically set as $1.0$. Note that during the target model training phase, we compute $\mathcal{L}$ on both sub-models and update them seperately.

\section{Experiments}

\subsection{Experimental Setup}
\textbf{Datasets.}
We evaluate our method on the main-stream UDA benchmark datasets: (1) \textbf{Office-31} dataset, which is a standard dataset for domain adaptation with three distinct domains: Amazon (\textbf{A}), 
Webcam (\textbf{W}), and DSLR (\textbf{D}). 
(2) \textbf{Office-Home} dataset, which is a more complex dataset with 15,500 images in 65 categories and four domains: Art (\textbf{Ar}), Clip Art (\textbf{Cl}), Product (\textbf{Pr}) and Real World (\textbf{Rw}). 
(3) \textbf{VisDA-2017} dataset, which contains two extremely different domains: \textbf{Synthetic}: Images collected from 3D rendering models; and \textbf{Real}: Real-world images. (4) \textbf{Digits} datasets, which contain three digits datasets:MNIST (\textbf{M}), USPS (\textbf{U}), and SVHN (\textbf{S}). The images in \textbf{M}, and \textbf{U} are gray-scale, while the images in \textbf{S} are colored. Extra experiments on \textbf{DomainNet} can be found in the supplementary material due to the space limitation.
\vspace{5pt}

\begin{table}[t]
    \centering
    \caption{Comparison of clean and robust accuracy \% of different methods on \textbf{Office-Home} and \textbf{VisDA-2017}. Note that we only show the average results here due to the space limitation. Please refer to the supplementary material for more detailed results.}
    \label{tab:home-vis}
        \fontsize{8}{12}\selectfont
        \begin{tabular}{ccccc}
            \toprule
            \multirow{2}{*}{Methods} 
            & \multicolumn{2}{c}{\textbf{Office-Home}} & \multicolumn{2}{c}{\textbf{VisDA-2017}} \\
             
            \cmidrule{2-5} 
            &Clean&Rob. &Clean&Rob. \\
            \midrule
            
            Baseline(UDA)           & \textbf{69.07} & 12.37 & 72.11 & 1.31    \\ \midrule
            Source-only             & 39.38 & 21.46 & 15.85 & 5.46    \\ \midrule
            AT+UDA                  & 46.99 & 30.40 & 66.77 & 18.61   \\ \midrule
            UDA+AT                  & 61.40 & 43.03 & 69.83 & 34.71   \\ \midrule
            ARTUDA\cite{lo2022exploring}\dag
                                    & -  & -  & 65.50 & 44.30   \\ \midrule
            RFA\cite{awais2021adversarial}   \dag   
                                    & 59.10 & 42.80 & 65.18 & 41.67   \\ \midrule
            Ours                    & 63.59 & \textbf{55.24} & \textbf{72.45} & \textbf{53.82}   \\ \bottomrule
        \end{tabular}
\end{table}

\noindent\textbf{Compared Methods.}
We compare our method with four baselines: (1) UDA baseline: Conventional UDA method that does not take robustness into account; (2) Source-only: the model is adversarially trained on the source data; (3) AT+UDA: The source data is first converted to adversarial examples, and then UDA is performed by aligning the adversarial source data and the clean target data. (4) UDA+AT: Naive self-training pipeline where the target model is adversarially trained on target data with pseudo labels generated from a pre-trained source model. We also compare with existing state-of-the-art RoUDA methods, including RFA~\cite{awais2021adversarial}, SRoUDA~\cite{zhu2022srouda}, ARTUDA~\cite{lo2022exploring}, and ASFR~\cite{10.1145/3503161.3548323} by using their reported results.
\vspace{5pt}

\noindent\textbf{Implementation Details.}
Following previous works~\cite{zhu2022srouda,lo2022exploring}, we use ResNet-50 as the backbone on \textbf{Office-31}, \textbf{Office-Home}, and \textbf{VisDA-2017} datasets, and DTN on \textbf{Digits} dataset, respectively. We adopt the basic UDA approach CDAN\cite{long2018conditional} as the UDA baseline and pretrain the clean source model. During the pre-training, we use the UDA code base TLlib\footnote{https://github.com/thuml/Transfer-Learning-Library} with the default settings of learning rate $0.001$ and epochs $20$. For the stage of training the robust target model, we use \(k=10, \epsilon=8/255\) in PGD-$k$ to generate adversarial examples and adjust the learning rate to $0.004$. The hyper-parameter $\alpha$ and $\beta$ are both set as $0.01$. The robust target model is trained by $70$ epochs with $100$ iterations per epoch. We use the model classification accuracy against PGD-$20$ with $\epsilon=8/255$ as the robust accuracy.

\subsection{Main Results}

\begin{table}[t]
    \centering
    \caption{Comparison of clean and robust accuracy \% of different methods on \textbf{Digits} dataset.}
    \label{tab:digits}
    \tabcolsep=2.2pt
    \fontsize{8}{12}\selectfont
        \begin{tabular}{ccccccccc}
            \toprule
            \multirow{2}{*}{Methods} 
            & \multicolumn{2}{c}{\textbf{M} $\xrightarrow{}$ \textbf{U}} & \multicolumn{2}{c}{\textbf{U} $\xrightarrow{}$ \textbf{M}} 
            & \multicolumn{2}{c}{\textbf{S} $\xrightarrow{}$ \textbf{M}} & \multicolumn{2}{c}{\textbf{Avg.}}\\
             
            \cmidrule{2-9} 
            &Clean&Rob. &Clean&Rob. &Clean&Rob. &Clean&Rob. \\
            \midrule
            
            Baseline(UDA) & 95.90 & 9.90  & \textbf{98.50} & 20.00 & 74.46 & 68.00  & 89.62 & 32.63  \\ \midrule
            Source-only   & 76.13 & 57.90 & 52.38 & 45.72 & 26.13 & 14.26 & 51.55 & 39.29   \\ \midrule
            AT+UDA        & 95.02 & 72.59 & 95.96 & 76.95 & 17.30 & 10.09  & 69.43 & 53.21  \\ \midrule
            UDA+AT        & 91.29 & 94.38 & 84.53 & 97.95 & 74.08 & 72.92  & 83.30 & 88.42  \\ \midrule
            SRoUDA\cite{zhu2022srouda} \dag       & 95.02 & 87.59 & \textbf{98.50} & 96.44 & 88.72 & 87.16  & 94.08 & 90.40  \\ \midrule
            Ours          & \textbf{98.15} & \textbf{95.27} & 98.21 & \textbf{97.72} & \textbf{92.98} & \textbf{92.38}  & \textbf{96.45} & \textbf{95.12}  \\ \bottomrule
        \end{tabular}
    
\end{table}

\begin{table*}[htbp]
    \centering
    \caption{Comparison of robust accuracy $\%$ of different UDA models produced by different methods against various common corruptions.}\label{tab:common_noise1}
     \fontsize{8}{9}\selectfont
    \begin{tabular}{c|cccccccccccc}
       \toprule
     
        Methods &\makecell{Gaussian\\ noise} &\makecell{Shot\\ noise} &\makecell{Motion\\ blur} &\makecell{Zoom\\ blur} &\makecell{Snow\\ noise} &\makecell{Brightness\\ noise} &\makecell{Translate\\ noise} &\makecell{Rotation\\ noise} &\makecell{Scale\\ noise} &\makecell{Speckle\\ noise} &\makecell{Gaussian\\ blur} &\makecell{Shear\\ noise}\\        \midrule
        Baseline (UDA) &79.24 &38.99 &90.19 &91.82 &66.03 &87.42 &91.71 &91.82	&87.42	&90.56	&52.83	&90.57\\
        \midrule   
        
        AT+UDA &62.89	&49.69	&64.15	&66.03	&62.26	&62.89	&65.40	&61.01	&40.88	&65.41	&40.25	&62.89 \\ \midrule

        UDA+AT &89.93	&88.05	&89.93	&90.56	&89.30	&89.93	&89.93	&89.31	&81.13	&89.93	&72.33	&87.42 \\ \midrule
        
        Ours &\textbf{94.34}	&\textbf{93.71}	&\textbf{91.19}	&\textbf{94.34}	&\textbf{91.82}	&\textbf{91.80}	&\textbf{93.71}	&\textbf{93.71}	&\textbf{89.30}	&\textbf{93.82}	&\textbf{86.34}	&\textbf{93.71} \\ \bottomrule
    \end{tabular}
\end{table*}

We first report the main results on \textbf{Office-31}, \textbf{Office-Home}, \textbf{VisDA-2017}, and \textbf{Digits} summarized in Tables\,\ref{tab:office-31}-\ref{tab:against-multy-attacks}. We can have the following observations:
\vspace{5pt}

\noindent\textbf{On Robustness and Discrimination.} From Tables\,\ref{tab:office-31}-\ref{tab:digits}, we can clearly see that the proposed method can effectively improve the adversarial robustness of UDA models, surpassing the baselines by a large margin. Specifically, we improve the model robustness from $3.21 \%$ to $84.28 \%$ on average for \textbf{Office-31}; $12.37\%$ to $55.24 \%$ on average for \textbf{Office-Home}; $1.31 \%$ to $53.82 \%$ on average for \textbf{VisDa-2017}; $32.63 \%$ to $95.12 \%$ on average for \textbf{Digits}, which largely closes the performance gap between UDA baseline with RoUDA. Moreover, when tested on small-scale datasets such as \textbf{Digits}, our method can achieve both near-optimal accuracy and robustness simultaneously on the target domain, \textit{i.e.}, $96.45\%$ clean and $95.12\%$ robustness accuracy on average. For comparison with other RoUDA methods, our method can still outperform the SOTA methods, \textit{e.g.}, the average model robustness improves $8.49\%$ over the second best method SRoUDA on \textbf{Office-31}, $12.15\%$ over the second best method RFA on \textbf{VisDa-2017}.  Besides, although the baselines and the SOTA RoUDA methods can improve the robustness compared to the UDA baseline, all of them show some decrease in clean accuracy, e.g., $88.26 \%$ to $84.21 \%$ on average for RFA on \textbf{Office-31}; and $72.11 \%$ to $65.50 \%$ on average for ARTUDA on \textbf{VisDa-2017}. On the contrary, our method can improve the model robustness without harming accuracy and even improve the clean accuracy over the UDA baseline on some tasks, \textit{e.g.}, $89.62 \%$ to $96.45 \%$ on average for \textbf{Digits}. These encouraging results validate that our MIRoUDA can effectively improve the adversarial robustness of UDA models and perform generally well on different domain adaptation tasks.

\begin{table}
   \centering
   \caption{Model performance \% comparison against different adversarial attacks on \textbf{VisDA-2017}.}
   \label{tab:against-multy-attacks}
   \tabcolsep=3pt
   \fontsize{8}{12}\selectfont
       \begin{tabular}{ccccccc}
           \toprule
           \multirow{2}{*}{Methods} & \multirow{2}{*}{Clean} & \multirow{2}{*}{FGSM\cite{goodfellow2014explaining}}  
            
           & \multicolumn{2}{c}{PGD-$k$\cite{madry2018towards}} & \multirow{2}{*}{\(CW_\infty\)\cite{cw_Carlini}} 
           &\multirow{2}{*}{AA\cite{AA_Croce}} \\
           \cmidrule{4-5}
           & &  &20 &100 \\
           \midrule
    
           Baseline(UDA) & 72.11 & 32.70  & 1.31  & 0.90  & 1.26  &0.49 \\ \midrule
           AT+UDA   & 66.77 & 36.04 & 18.61 & 17.60 & 17.92 &16.31 \\ \midrule
           UDA+AT        & 69.83 & 53.62  & 34.71 & 34.37 & 34.68 &29.84 \\ \midrule
           Ours          & \textbf{72.45} & \textbf{68.29}  & \textbf{53.82} & \textbf{53.66} & \textbf{53.68} &\textbf{52.33} \\ \bottomrule
       \end{tabular}
\end{table}

\begin{figure}
   \centering
   \includegraphics[width=1.0\linewidth]{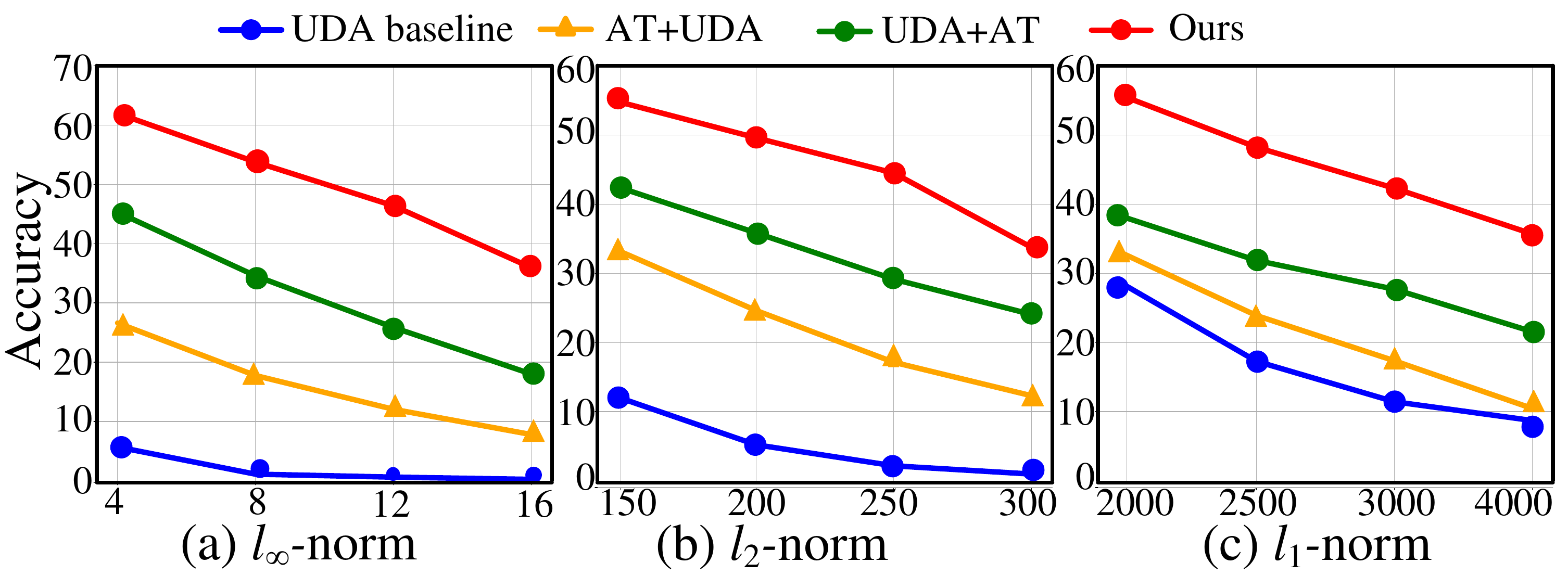}
   \caption{Robust accuracy(\%) of ResNet50 trained with $l_\infty$ of $\epsilon=8/255$ boundary against unseen attacks. For unseen attacks, we use PGD-50 under different-sized $l_\infty$ balls, and other types of norm balls, \textit{e.g.}, $l_1$, and $l_2$.}
    \vspace{-10pt}
   \label{fig:multi-attacks}
\end{figure}

\vspace{5pt}
\noindent\textbf{On Generalization.} Here we test the model robustness generalization from three aspects: 1) \textit{Against different adversarial attack strategies}; 2) \textit{Against unseen adversaries}; and 3) \textit{Against common noises}. Specifically, we first evaluate the model generalization on \textbf{VisDA-2017} to different attacks, including FGSM~\cite{goodfellow2014explaining}, PGD-$k$~\cite{madry2018towards}, CW$_\infty$~\cite{cw_Carlini}, and AutoAttack (AA)~\cite{AA_Croce}, shown in Table\,\ref{tab:against-multy-attacks}. One can see that our method achieves the best robustness against all the attacks, surpassing the baseline around $25\%$ on the strongest attack AA. We then conduct the experiments on the model generalization to unseen adversaries, \textit{i.e.}, robustness against different threat norm constraints (\textit{e.g.}, $l_1$, and $l_2$) and radius $\epsilon$. As can be seen in Figure\,\ref{fig:multi-attacks}, by incorporating the MI theory and the dual-model architecture, our proposed method can consistently and significantly improve the robustness against all unseen adversaries. All these results indicate the superior generalization ability of our proposed method to various and multiple perturbations.

Next, we further test the model generalization to the common noises on task \textbf{A} $\rightarrow$ \textbf{W}, where we add common corruptions instead of adversarial perturbations on the target data according to~\cite{hendrycks2019robustness}. Please refer to Sec.\,\textbf{A.2} in the supplementary material for the details about noise addition. The results are reported in Table\,\ref{tab:common_noise1}, where our method shows excellent robust generalization ability to the common noises, especially on the powerful corruptions, \textit{e.g.}, our method improves from $38.99 \%$ to $93.71 \%$ against Shot noise over the UDA baseline, and $52.83 \%$ to $86.34 \%$ against Gaussian blur. In addition, our method can achieve over $90\%$ accuracy against most of the corruptions, revealing the high practical utility of the proposed method.

\begin{table}
    \centering
    \caption{Experimental results of component ablations on our MIRoUDA.}
    \label{tab:component-ablation}
    \fontsize{8.5}{12.5}\selectfont
        \begin{tabular}{cccccccccc}
            \toprule
            \multicolumn{3}{c}{\(\mathcal{L}_\textbf{MI}\)}
            &\makebox[0.05\linewidth]{\multirow{2}{*}{\(\mathcal{L}_{cs}\)}} 
            &\multicolumn{2}{c}{\textbf{A} \(\xrightarrow{}\) \textbf{W}} 
            & \multicolumn{2}{c}{\textbf{A} \(\xrightarrow{}\) \textbf{D}} \\
             
            \cmidrule{5-8} 
            
            ($\mathcal{L}_{dis}$, &$\mathcal{L}_{rob}$,&$\mathcal{L}_{gen}$)&&Clean&Rob. &Clean&Rob. \\
            \midrule
            
            \(\checkmark\)&&&& 88.67 & 76.73 & 85.00 & 58.00   \\ \midrule
            \(\checkmark\)&\(\checkmark\)&&& 79.75 & 88.68 & 65.00 & 72.00  \\ \midrule
            \(\checkmark\)&\(\checkmark\)&\(\checkmark\) &&89.31 & 88.68 & 90.00 & 76.00   \\ \midrule
            

            \(\checkmark\)&\(\checkmark\)&&\(\checkmark\)&93.71 & 90.82 & 87.00 & 76.00   \\ \midrule
            \(\checkmark\)&\(\checkmark\)&\(\checkmark\)&\(\checkmark\)& 94.34 & 91.19 & 93.00 & 83.00   \\ 
            \bottomrule
        \end{tabular}
\end{table}

\subsection{Further Analysis}\label{sec:analysis}
We further analyze our MIRoUDA from four perspectives: 1) Component ablations; 2) Sensitivity to different UDA baselines in the pre-training; 3) Sensitivity to hyper-parameters; and 4) Feature space visualization.

\noindent\textbf{Component Ablations.}
We examine the effectiveness of each component in our method on \textbf{A}\(\rightarrow\) \textbf{W} and \textbf{A}\(\rightarrow\) \textbf{D} tasks in the \textbf{Office-31} dataset following the previous settings. Table\,\ref{tab:component-ablation} shows that only using $\mathcal{L}_{dis}$ can improve the model robustness to some extent. Adding the $\mathcal{L}_{rob}$ can significantly improve the robustness, but takes the clean accuracy as a sacrifice. When combing all the $\mathcal{L}_{dis}$, $\mathcal{L}_{rob}$, and $\mathcal{L}_{gen}$, both the clean and robust accuracy improve a lot. We can also see that without $\mathcal{L}_{gen}$ but adding the $\mathcal{L}_{cs}$ can improve the clean accuracy as well, indicating its strong ability in clean accuracy preservation. In general, the full MIRoUDA performs the best clean accuracy and robustness, where it achieves the highest values in terms of all the metrics. 

\vspace{3pt}

\noindent\textbf{Sensitivity to Different UDA Baselines in Pre-training.}
We also test the model performance when using different UDA baselines in the source model pre-training. Here, we use DAN~\cite{long2015learning}, DANN~\cite{ganin2015unsupervised}, BSP~\cite{chen2019transferability}, MDD~\cite{MDD_ICML_19} and SDAT~\cite{sdat_rangwani2022closer} in place of CDAN in source model pre-training of MIRoUDA and test on \textbf{A}\(\rightarrow\) \textbf{W} and \textbf{W}\(\rightarrow\)\textbf{A} tasks following the previous settings. The results are reported in Table \ref{tab:different-uda}. We can find that (i) using our MIRoUDA can significantly improve the model robustness of all the UDA baselines without harming the clean accuracy a lot; (ii) Both the clean and robust accuracy vary as the change of UDA baselines in the source model pre-training, which once again indicates that the model pre-training plays a crucial role in self-training pipelines. A better UDA baseline can help produce more robust target models.

\begin{table}
    \centering
    \caption{Clean/Robust accuracy \% comparison of using different UDA baselines in source model pre-training.}
    \label{tab:different-uda}
    \tabcolsep=5pt
    \fontsize{8}{12}\selectfont
        \begin{tabular}{ccccc}
            \toprule
            \multirow{2}{*}{Methods} 
            & \multicolumn{2}{c}{\textbf{A} \(\xrightarrow{}\) \textbf{W}} & \multicolumn{2}{c}{\textbf{W} \(\xrightarrow{}\) \textbf{A}} \\
             
            \cmidrule{2-5} 
            &Baseline&Ours &Baseline&Ours \\
            \midrule
            DAN~\cite{long2015learning}         &84.15/1.51 & 81.76/77.36 &64.61/10.76 &67.55/60.11   \\ \midrule
            DANN~\cite{ganin2015unsupervised}   &91.19/2.78 & 93.71/89.31 &73.27/20.98 &73.76/65.96   \\ \midrule
            CDAN~\cite{long2018conditional}     &91.19/0.00 & 94.34/91.19 &73.30/11.96 &70.92/65.10   \\ \midrule
            BSP~\cite{chen2019transferability}  &92.70/1.76 & 93.08/91.82 &74.15/17.47 &74.82/67.55   \\ \midrule
            MDD~\cite{MDD_ICML_19}              &93.08/2.26 & 92.45/88.69 &72.35/19.28 &74.11/66.49   \\
            \midrule
            SDAT~\cite{sdat_rangwani2022closer} &93.71/1.79 & 93.71/90.57 &75.15/10.34  &75.89/68.62   \\ 
            \bottomrule
        \end{tabular}
\end{table}

\begin{figure}
  \centering
  \includegraphics[width=0.45\textwidth]{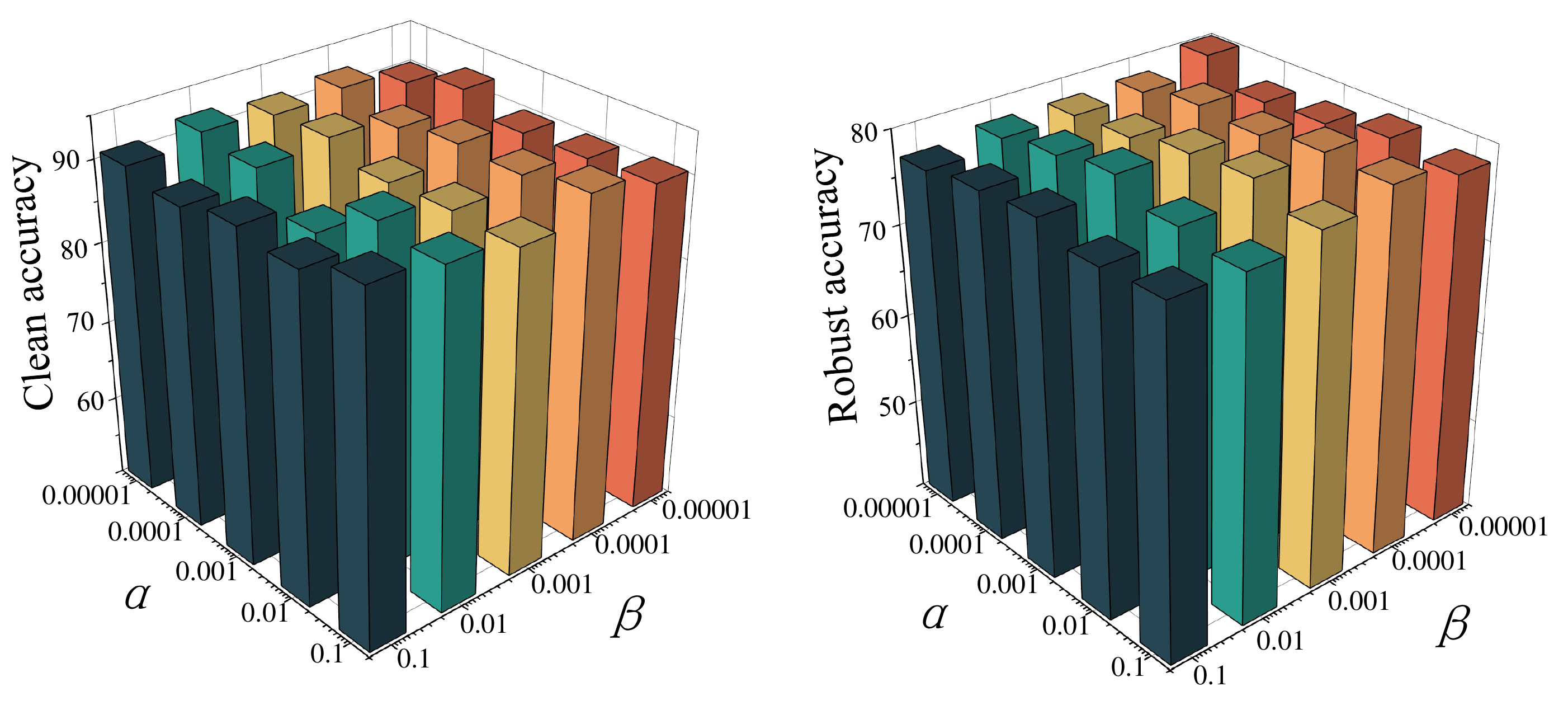}
  \caption{The effect of hyper-parameters $\alpha$ and $\beta$ on model performance in clean accuracy (left) and robust accuracy (right). The stable performance shows its robustness against hyper-parameters.}
  \label{fig:para}
\end{figure}

\begin{figure}
  \centering
  \includegraphics[width=0.48\textwidth]{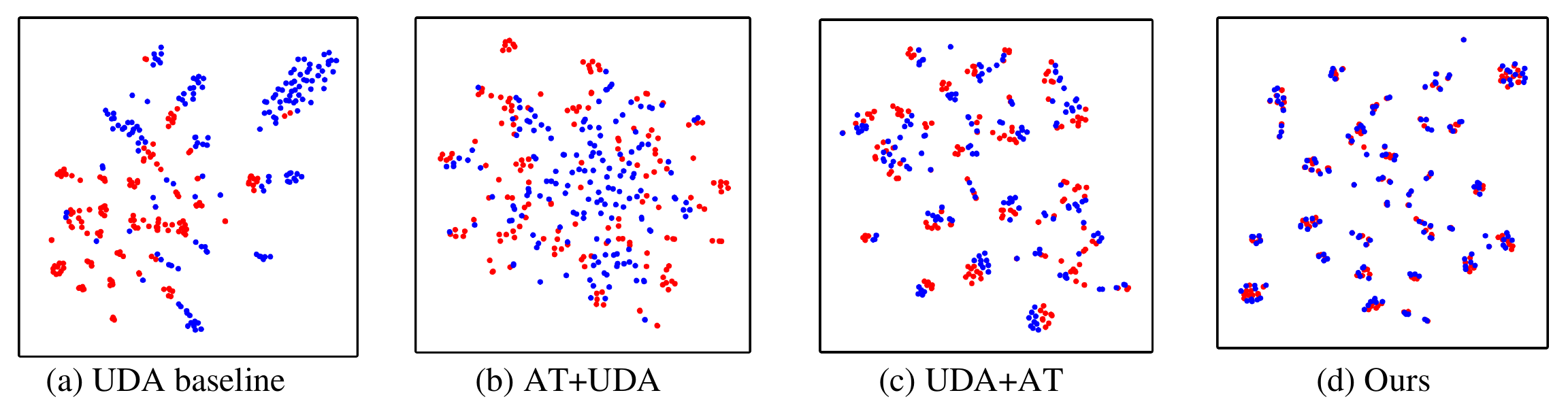}
  \caption{(a)-(d): Visualization of features using t-SNE embeddings~\cite{NIPS2002_6150ccc6} from models trained with different methods on A\(\rightarrow\)W task. The \textcolor{red}{red} dots stand for clean target data, and the \textcolor{blue}{blue} dots stand for adversarial examples of target data.}
   \vspace{-5pt}
  \label{fig:tsne}
\end{figure}
\vspace{3pt}
\noindent\textbf{Sensitivity to Hyper-parameters.}
We investigate the influence of the hyper-parameters $\alpha$ and $\beta$ on task \textbf{A}\(\rightarrow\)\textbf{D} following the previous settings. As shown in Figure\,\ref{fig:para}, the performance of MIRoUDA is stable under different values of $\alpha$ and $\beta$, which demonstrates its robustness against the hyper-parameters. However, when one of the $\mathcal{L}_{\textbf{MI}}$ is removed, the model encounters a significant drop as analyzed in the component ablation study. 
\vspace{3pt}

\noindent\textbf{Feature Space Visualization.}
Figure \ref{fig:tsne} provides a visual analysis of the representations produced by different methods on task \textbf{A}\(\rightarrow\)\textbf{W} using t-SNE embeddings~\cite{NIPS2002_6150ccc6}. The features of clean examples and adversarial examples are completely dispersed with the UDA baseline (see Figure \ref{fig:tsne} (a)), since it does not take robustness into account. Injecting AT can improve this situation, but it does not align well due to source domain bias and poor pseudo-label quality (see Figure \ref{fig:tsne} (b)-(c)). By contrast, our proposed MIRoUDA method better aligns the feature distributions and exhibits superior category discriminability. 

\section{Conclusion}

In this paper, we unify the RoUDA task under the mutual information theory and propose a novel theoretic-grounded method MIRoUDA accordingly. Specifically, we first theoretically show that the proposed MI optimization could achieve robust, discriminative, and generalized representation learning. We build the whole framework based on the self-training pipeline and then design a dual-model architecture for robust target model learning. With the help of an additional regularizer, our method demonstrates superior performance over baselines and state-of-the-arts on various benchmarks in the extensive experiments.


\bibliographystyle{ACM-Reference-Format}
\bibliography{reference}

\clearpage
\title{Supplementary Materials: Towards Trustworthy Unsupervised Domain Adaptation: A Representation Learning Perspective for Enhancing Robustness, Discrimination, and Generalization}

\section*{A More Experimental Details}
\label{sec:addi_ablation} 
In this section, we describe the experimental details that could not be included in the main paper due to the page limits.
\subsection*{A.1 Details in Hyper-parameter Analysis}
\begin{figure}[htbp]
   \centering
   \includegraphics[width=1.0\linewidth]{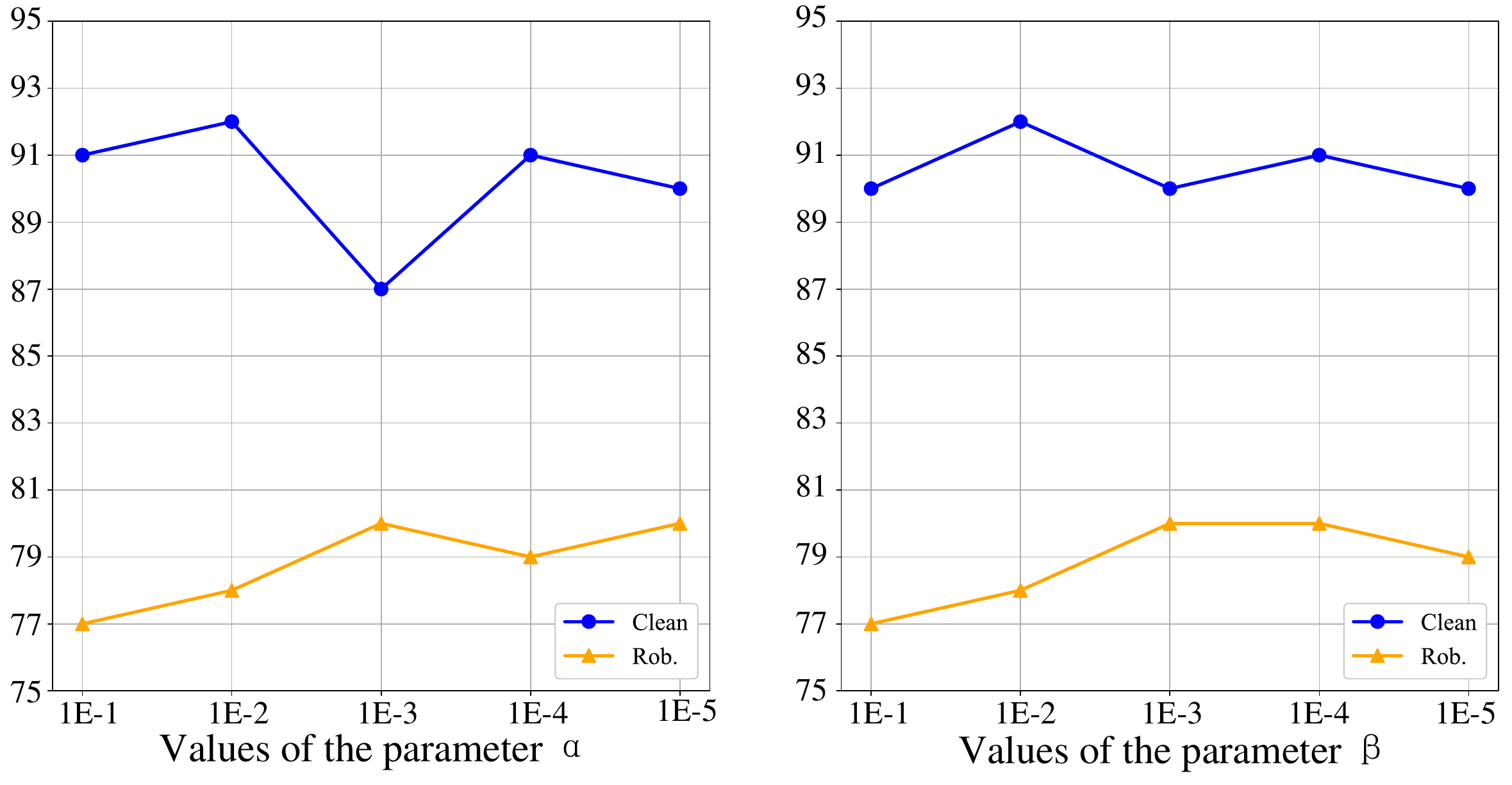}
   \caption{A close look at the model sensitivity to Hyper-parameter. We take the task \textbf{A}\(\rightarrow\)\textbf{D} as examples. The \textcolor{blue}{blue} dotted line denotes clean accuracy and \textcolor{orange}{orange} dotted line denotes the robustness evaluated with PGD-20, $\lambda$=8/255.}
   \label{fig:appandix_para}
\end{figure}
\noindent We only show the overall results of parameter analysis in our main paper and found that they are robust to the parameter changes. Here we take a closer look at the model changes to the parameters $\alpha$ and $\beta$ in the objective function of proposed MI constraints, as defined in Eq. \eqref{eq:rmi} which controls the strength of MI constraints in different levels. We conducted experiments on \textbf{A}\(\rightarrow\) \textbf{D} tasks with the UDA baseline clean accuracy 92.97\%, and present the results in Figure \ref{fig:para} for different $\alpha,\beta \in [1e-1,1e-2,1e-3,1e-4,1e-5]$. As can be seen in Figure\,\ref{fig:appandix_para}, the clean accuracy reaches the peak when the values of parameters of $\alpha$ and $\beta$ are set to 0.01. Although reducing the values of the parameters, e.g. decreasing $\alpha$ from 1e-2 to 1e-3 and $\beta$ from 1e-2 to 1e-5, can slightly improve the robustness, it results in a relative large decrease in clean accuracy. The size of parameters does not have a strong relationship with the robustness and clean accuracy, and the perfomance is quite stable.  We set $\alpha$=0.01 and $\beta$=0.01, which is good for most datasets.

\subsection*{A.2 Settings of Common Corruptions}
\begin{figure}[!t]
    \includegraphics[width=1.0\linewidth]{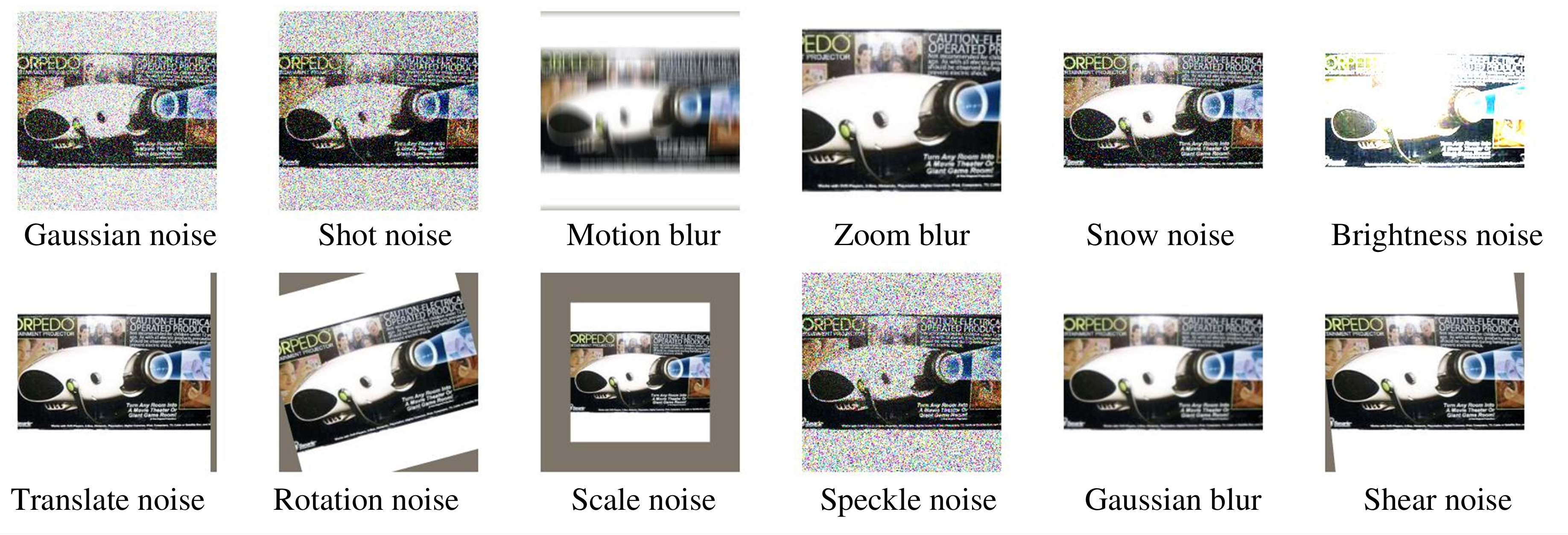}
    \caption{An example of image generated with the transformations of our common corruptions}
    \label{fig:noise}
\end{figure}

In this section, we describe the details of common corruptions addition in evaluating the performance of the model robustness against various common corruptions. Specifically, we add common corruptions on the target data, including Gaussian noise with $\sigma=\{0.02, 0.04, 0.06\}$, shot noise with $f=\{200, 400, 600\}$, motion blur with kernel size $5$, zoom blur with $f=\{1,2,...,31\}$, snow noise with intensity in $[0, 1]$ and brightness in $[0,1]$, brightness noise with $f=0.5$, translation with $f\in [-7, 9]$, rotation with angle $-180$, scaling with the sacle factor in $[0.7, 1.3]$, speckle noise with $r=\{0.05, 0.07, 0.1\}$, Gaussian blur with kernel size $21$, and shear noise with $r\in [-15, 15]$. An example of image generated with the transformations of our common corruptions is shown in Figure \ref{fig:noise}.

\begin{table*}[!t]
    \centering
    \caption{Comparison of accuracy and robustness \% of different method on \textbf{VisDA-2017} dataset.}
    \label{tab: detail_visda2017}
    \resizebox{\linewidth}{!}{
        \begin{tabular}{ccccccccccccccc}
            \toprule
              \multicolumn{2}{c}{\textbf{Methods}}
             & \textbf{plane} & \textbf{bicycle}  & \textbf{bus}        & \textbf{car} 
             & \textbf{horse} & \textbf{knife}    & \textbf{motorc}     & \textbf{person}
             & \textbf{plant} & \textbf{sktboard} & \textbf{train}      & \textbf{truck}     & \textbf{Avg.} \\ 

            \midrule
            Baseline(UDA)
                &Clean  & 94.76 &69.27 &85.69 &50.80 &88.98 &19.08  &88.27 &76.90 &85.38 &87.86  &86.40  &44.43  &72.11 \\ 
                &Rob.   & 5.60  &0.36  &0.43  &0.04  &2.64  &6.51   &0.02  &0.33  &1.74  &0.00   &3.12   &0.02   &1.31  \\ \midrule
            Source-only
                &Clean  & 7.54  &0.26  &65.80 &10.53 &0     &0      &45.77 &0     &6.37  &43.82  &10.14  &0      &15.85 \\ 
                &Rob.   & 9.36  &0.39  &17.30 &20.29 &0     &0      &0.52  &0     &11.44 &0.89   &5.85   &0      &5.46  \\ \midrule
            AT+UDA
                &Clean  & 87.85 &67.02 &74.58 &38.45 &75.48 &63.95  &86.51 &68.63 &79.91 &76.46  &80.08  &46.09  &66.77 \\ 
                &Rob.   & 51.81 &4.89  &10.04 &17.83 &23.17 &3.95   &25.98 &1.15  &16.71 &9.29   &48.28  &3.37   &18.61 \\ \midrule
            UDA+AT
                &Clean  & 95.22 &61.21 &84.23 &55.48 &75.65 &13.51  &89.65 &77.96 &86.76 &78.19  &82.19  &36.00  &69.83 \\ 
                &Rob.   & 76.51 &49.37 &51.92 &18.03 &60.61 &37.66  &8.11  &36.05 &30.46 &39.96  &48.11  &10.89  &34.71 \\ \midrule
            ARTUA\cite{lo2022exploring}
                &Clean  & -- &-- &-- &-- &-- & -- &-- &-- &-- &-- &--  &--  &65.50 \\ 
                &Rob.   & -- &-- &-- &-- &-- & -- &-- &-- &-- &--  &--  &--  &44.30 \\ \midrule
            RFA\cite{awais2021adversarial}
                &Clean  & -- &-- &-- &-- &-- & -- &-- &-- &-- &--  &--  &--  &65.18 \\ 
                &Rob.   & -- &-- &-- &-- &-- & -- &-- &-- &-- &--  &--  &--  &41.67 \\ \midrule
            Ours
                &Clean  & 92.46 &68.46 &84.36 &49.64 &91.03 &22.74  &90.11 &74.91 &86.27 &84.21  &84.04  &53.87  &\textbf{72.45} \\ 
                &Rob.   & 80.72 &43.40 &68.00 &40.48 &71.50 &14.89  &71.28 &50.06 &66.29 &66.32  &61.35  &20.44  &\textbf{53.82} \\ \midrule            
        \end{tabular}
    }
\end{table*}

\begin{table*}[!t]
        \tabcolsep=3.2pt
        \fontsize{8}{12}\selectfont
        \caption{Comparison of clean and robust accuracy $\%$ of different UDA models produced by different methods on \textbf{Office-Home} dataset.}
        \label{tab: detail_home}
        \begin{tabular}{ccccccccccccccc}
            \toprule
              \multicolumn{2}{c}{\textbf{Methods}}
             & \textbf{Ar}\(\xrightarrow{}\)\textbf{Cl} & \textbf{Ar}\(\xrightarrow{}\)\textbf{Pr} & \textbf{Ar}\(\xrightarrow{}\)\textbf{Rw}
             & \textbf{Cl}\(\xrightarrow{}\)\textbf{Ar} & \textbf{Cl}\(\xrightarrow{}\)\textbf{Pr} & \textbf{Cl}\(\xrightarrow{}\)\textbf{Rw}
             & \textbf{Pr}\(\xrightarrow{}\)\textbf{Ar} & \textbf{Pr}\(\xrightarrow{}\)\textbf{Cl} & \textbf{Pr}\(\xrightarrow{}\)\textbf{Rw}
             & \textbf{Rw}\(\xrightarrow{}\)\textbf{Ar} & \textbf{Rw}\(\xrightarrow{}\)\textbf{Cl} & \textbf{Rw}\(\xrightarrow{}\)\textbf{Pr}
             & \textbf{Avg.} \\ 

            \midrule

            Baseline(UDA)
                &Clean  & 56.95 &72.79 &77.39 &62.09 &71.86 &71.08  &60.40 &55.97 &80.12 &74.00  &62.57  &83.60  &69.07 \\ 
                &Rob.   & 19.68 &13.38 &7.41  &4.21  &16.31 &7.64   &4.86  &21.03 &9.43  &4.94   &21.65  &17.91  &12.37 \\ \midrule
            Source-only
                &Clean  & 28.67 &42.66 &50.21 &24.36 &43.92 &39.30  &22.56 &28.81 &47.97 &44.62  &39.30  &60.14  &39.38 \\ 
                &Rob.   & 26.85 &22.10 &25.59 &5.13  &20.56 &14.27  &9.74  &19.30 &23.08 &18.46  &31.19  &41.26  &21.46 \\ \midrule
            AT+UDA
                &Clean  & 34.87 &39.24 &50.91 &44.17 &48.41 &54.95  &38.31 &32.37 &58.87 &55.33  &43.73  &62.74  &46.99 \\ 
                &Rob.   & 31.07 &28.20 &36.03 &16.44 &33.25 &28.55  &14.91 &29.53 &33.60 &28.47  &39.34  &46.38  &30.40 \\ \midrule
            UDA+AT
                &Clean  & 53.95 &71.06 &67.89 &46.30 &71.51 &62.96  &51.03 &51.32 &71.10 &50.41  &58.89  &80.41  &61.40 \\ 
                &Rob.   & 45.70 &58.45 &38.07 &28.19 &56.42 &41.28  &27.98 &42.27 &37.96 &27.16  &46.85  &65.99  &43.03 \\ \midrule
            RFA\cite{awais2021adversarial}
                &Clean  & -- &-- &-- &-- &-- & -- &-- &-- &-- &--  &--  &--  &59.10 \\ 
                &Rob.   & -- &-- &-- &-- &-- & -- &-- &-- &-- &--  &--  &--  &42.80 \\ \midrule
            Ours
                &Clean  & 55.44 &71.28 &70.76 &53.29 &71.96 &65.25  &50.82 &52.35 &70.09 &60.49  &60.48  &80.86  &63.59 \\ 
                &Rob.   & 53.60 &67.00 &56.54 &39.51 &67.00 &53.55  &37.86 &50.29 &56.54 &47.33  &57.85  &75.79  &55.24 \\ \midrule
    \end{tabular}
\end{table*}

\begin{table}[!t]
    \centering
    \caption{Replacing components of RoUDA on \textbf{Office-31} \textbf{A}\(\rightarrow\)\textbf{W} and \textbf{W}\(\rightarrow\)\textbf{A} tasks. }
    \tabcolsep=4pt
    \label{tab:replacing-components}
    \fontsize{8}{12}\selectfont
        \begin{tabular}{ccccc}
            \toprule
            \multirow{2}{*}{Methods} 
            & \multicolumn{2}{c}{\textbf{A} \(\xrightarrow{}\) \textbf{W}} & \multicolumn{2}{c}{\textbf{W} \(\xrightarrow{}\) \textbf{A}} \\
             
            \cmidrule{2-5} 
            &Clean&Rob. &Clean&Rob. \\
            \midrule
            
            UDA+AT                          & 88.67 & 76.73 & 69.33 & 60.28   \\ \midrule
            UDA+TRADES                      & 87.42 & 79.25 & 66.49 & 62.94   \\ \midrule
            Ours (TRADES)                    & \textbf{94.34} & 91.82 & 68.99 & 64.72   \\ \midrule
            Ours (Consensus on AEs)       & 93.08 & 91.82 & 69.39 & 64.25   \\ \midrule
            Ours        & \textbf{94.34} & \textbf{93.08} & \textbf{70.92} & \textbf{65.10}   \\ \bottomrule
        \end{tabular}
\end{table}

\section*{B Detailed Results}
\subsection*{B.1 Detailed Results on VisDA-2017}
The detailed results on \textbf{VisDA-2017} are shown in Table \ref{tab: detail_visda2017}. The proposed method achieves an average clean accuracy of 72.45\% and robustness of 53.82\%, outperforming the existing SOTA methods. In particular, for the hard category, such as knife, UDA+AT even leads to higher robust accuracy than clean accuracy. The reason for this phenomenon may be the notably low pseudo-label accuracy of only 19.08\%, indicating that the model is severely affected by adversarial perturbations. However, our method addresses this issue, achieving a clean accuracy of 22.74\% and robustness of 14.89\%.

\subsection*{B.2 Detailed Results on Office-Home}
In this part, we show the detailed results on \textbf{Office-Home} in Table \ref{tab: detail_home}. As can be seen, AT+UDA aligned the target data with the adversarial examples only generated in source data, which results in a dramatically decrease in clean accuracy and thus poorer robustness. The proposed MIRoUDA achieves an average accuracy of 63.59\% and robustness of 55.24\%, outperforming the existing SOTA methods.

\section*{C Extra Experiments}
\subsection*{C.1 Experiments on DomainNet}
We further evaluate the performance of the proposed method on \textbf{DomainNet}~\cite{domainnet_peng2019moment}, which contains over 0.6 million images across 345 categories with 6 subdomains. Here, we conduct experiments on 4 subdomains (Clipart, Painting, Real, Sketch), which consist of 12 sub-experiments. As the results shown in table \ref{tab: detail_domainnet}, our method achieves the state-of-the-art model robustness without harming clean accuracy compared to other schemes for improving UDA robustness. Such promising results provide evidence of the superiority of our MIRoUDA for processing complex datasets.

\begin{table*}
        \centering
        \tabcolsep=7pt
        \fontsize{8}{12}\selectfont
        \caption{Comparison of clean and robust accuracy $\%$ of different UDA models produced by different methods on \textbf{DomainNet} dataset.}
        \label{tab: detail_domainnet}
        \begin{tabular}{ccccccccccccccc}
            \toprule
              \multicolumn{2}{c}{{\textbf{Methods}}}
             & \textbf{c}\(\xrightarrow{}\)\textbf{p} & \textbf{c}\(\xrightarrow{}\)\textbf{r} & \textbf{c}\(\xrightarrow{}\)\textbf{s}
             & \textbf{p}\(\xrightarrow{}\)\textbf{c} & \textbf{p}\(\xrightarrow{}\)\textbf{r} & \textbf{p}\(\xrightarrow{}\)\textbf{s}
             & \textbf{r}\(\xrightarrow{}\)\textbf{c} & \textbf{r}\(\xrightarrow{}\)\textbf{p} & \textbf{r}\(\xrightarrow{}\)\textbf{s}
             & \textbf{s}\(\xrightarrow{}\)\textbf{c} & \textbf{s}\(\xrightarrow{}\)\textbf{p} & \textbf{s}\(\xrightarrow{}\)\textbf{r}
             & \textbf{Avg.} \\ 

            \midrule

            Baseline(UDA)
                &Clean  & 41.29 &56.28 &48.00 &46.11 &58.21 &40.05  &55.80 &53.50 &42.76 &58.70 &46.40 &55.65 &47.23 \\ 
                &Rob.   & 3.22  &5.87  &15.00 &14.56 &5.91  &10.34  &14.30 &2.40  &10.03 &24.10 &3.17  &6.20  &9.59  \\ \midrule
            Source-only
                &Clean  & 25.42 &36.35 &30.22 &18.03 &30.19 &16.32  &27.38 &36.87 &19.96 &31.54 &29.47 &33.82 &27.96 \\ 
                &Rob.   & 10.28 &16.78 &21.81 &20.07 &18.91 &13.53  &23.85 &13.71 &13.07 &24.42 &11.79 &15.23 &16.95 \\ \midrule
            AT+UDA
                &Clean  & 27.50 &40.62 &32.01 &26.39 &36.09 &26.98  &35.82 &34.85 &26.71 &39.46 &34.64 &40.62 &33.47 \\ 
                &Rob.   & 13.37 &20.40 &23.58 &23.13 &22.77 &22.16  &30.12 &18.72 &16.85 &31.40 &14.41 &21.43 &21.53 \\ \midrule
            UDA+AT
                &Clean  & 34.82 &48.01 &30.31 &38.22 &51.48 &30.74  &49.77 &43.75 &35.89 &52.28 &40.70 &52.89 &42.41 \\ 
                &Rob.   & 16.80 &25.31 &31.69 &35.07 &26.97 &30.29  &41.43 &23.39 &30.43 &40.01 &23.77 &27.84 &29.42 \\ \midrule
            Ours
                &Clean  & 40.25 &53.16 &42.39 &44.32 &53.63 &36.53  &52.99 &48.64 &37.45 &56.74 &44.94 &54.04 &47.09 \\ 
                &Rob.   & 20.84 &31.55 &32.69 &37.38 &29.32 &34.02  &42.53 &24.31 &31.21 &44.27 &24.01 &28.14 &31.69 \\ \midrule
    \end{tabular}
\end{table*}

\subsection*{C.2 Replacing the Component of MIRoUDA}
We conducted an evaluation of the robustness and accuracy of UDA models on \textbf{Office-31} \textbf{A}\(\rightarrow\)\textbf{W} and \textbf{W}\(\rightarrow\)\textbf{A} tasks, with a focus on the impact of component replacement. We solely apply the proposed consensus regularizer in UDA+AT as it effectively showcases the effects of replacing different components. As shown in Table \ref{tab:replacing-components}, when combing TRADES with UDA (UDA+TRADES) instead of combing AT with UDA (UDA+AT), the final robustness increases while at the expense of a decrease in clean accuracy. We then further replace the AT with TRADES in our method, as shown in the fourth row of Table \ref{tab:replacing-components}, both clean accuracy and robustness decrease due to the noisy pseudo labels. We also investigate to force consensus on the adversarial example logits output in the target model and the clean data logits output in the source model. The results are presented in the fifth row of Table \ref{tab:replacing-components}. We can observe that consensus on the adversarial representation in the target model and clean representation in the source model can also achieve certain improvements over naive UDA+AT. But by comparison, the consensus on clean representations in our method can achieve the best results.


\end{document}